\documentclass{article}

\usepackage{microtype}
\usepackage{graphicx}
\usepackage{booktabs} % for professional tables

\usepackage{hyperref}

\usepackage{enumitem}
\usepackage{amsfonts}
\usepackage{amsmath, amsthm, amssymb}
\usepackage{color}
\usepackage{subcaption}
\usepackage{relsize}
\usepackage[usenames,dvipsnames,svgnames,table]{xcolor}
\usepackage{algorithm,algorithmic}
\newcommand{\bs}{\boldsymbol}
\newcommand{\bx}{\bs{x}}
\newcommand{\bz}{\bs{z}}
\newcommand{\by}{\bs{y}}

\newcommand{\tabincell}[2]{\begin{tabular}{@{}#1@{}}#2\end{tabular}}

\renewcommand{\algorithmicrequire}{\textbf{Input:}}  % Use Input in the format of Algorithm
\renewcommand{\algorithmicensure}{\textbf{Output:}} % Use Output in the format of Algorithm

\usepackage[accepted]{icml2018}
\icmltitlerunning{Path-Level Network Transformation for Efficient Architecture Search}

\begin{document}

\twocolumn[
\icmltitle{Path-Level Network Transformation for Efficient Architecture Search}



\icmlsetsymbol{equal}{*}

\begin{icmlauthorlist}
\icmlauthor{Han Cai}{SJTU}
\icmlauthor{Jiacheng Yang}{SJTU}
\icmlauthor{Weinan Zhang}{SJTU}
\icmlauthor{Song Han}{MIT}
\icmlauthor{Yong Yu}{SJTU}
\end{icmlauthorlist}

\icmlaffiliation{SJTU}{Shanghai Jiao Tong University, Shanghai, China}
\icmlaffiliation{MIT}{Massachusetts Institute of Technology, Cambridge, USA}

\icmlcorrespondingauthor{Han Cai}{hcai@apex.sjtu.edu.cn}

\icmlkeywords{efficient architecture search, path-level function-preserving transformation, reinforcement learning, tree-structured architecture space}

\vskip 0.3in
]



\printAffiliationsAndNotice{}  % leave blank if no need to mention equal contribution

\begin{abstract}
We introduce a new function-preserving transformation for efficient neural architecture search. This network transformation allows reusing previously trained networks and existing successful architectures that improves sample efficiency. We aim to address the limitation of current network transformation operations that can only perform layer-level architecture modifications, such as adding (pruning) filters or inserting (removing) a layer, which fails to change the topology of connection paths. Our proposed path-level transformation operations enable the meta-controller to modify the path topology of the given network while keeping the merits of reusing weights, and thus allow efficiently designing effective structures with complex path topologies like Inception models. We further propose a bidirectional tree-structured reinforcement learning meta-controller to explore a simple yet highly expressive tree-structured architecture space that can be viewed as a generalization of multi-branch architectures. We experimented on the image classification datasets with limited computational resources (about 200 GPU-hours), where we observed improved parameter efficiency and better test results (97.70\% test accuracy on CIFAR-10 with 14.3M parameters and 74.6\% top-1 accuracy on ImageNet in the mobile setting), demonstrating the effectiveness and transferability of our designed architectures.
\end{abstract}

\section{Introduction}
% architecture search (its limitation) -> function preserving architecture search (EAS)
Designing effective neural network architectures is crucial for the performance of deep learning. While many impressive results have been achieved through significant manual architecture engineering \cite{simonyan2014very,szegedy2015going,he2016deep,huang2016densely}, this process typically requires years of extensive investigation by human experts, which is not only expensive but also likely to be suboptimal. Therefore, automatic architecture design has recently drawn much attention \cite{zoph2016neural,zoph2017learning,liu2017progressive,cai2018efficient,real2018regularized,pham2018efficient}. 

Most of the current techniques focus on finding the optimal architecture in a designated search space starting from scratch while training each designed architecture on the real data (from random initialization) to get a validation performance to guide exploration. Though such methods have shown the ability to discover network structures that outperform human-designed architectures when vast computational resources are used, such as \citet{zoph2017learning} that employed 500 P100 GPUs across 4 days, they are also likely to fail to beat best human-designed architectures \cite{zoph2016neural,real2017large,liu2017hierarchical}, especially when the computational resources are restricted. Furthermore, insufficient training epochs during the architecture search process (much fewer epochs than normal to save time) may cause models to underperform \cite{baker2016designing}, which would harm the efficiency of the architecture search process. 

% limitation of current Net2Net Techniques, from Modern CNN Architecture perspective
Alternatively, some efforts have been made to explore the architecture space by network transformation, starting from an existing network trained on the target task and reusing its weights. For example, \citet{cai2018efficient} utilized Net2Net \cite{chen2015net2net} operations, a class of function-preserving transformation operations, to further find high-performance architectures based on a given network, while \citet{ashok2017n2n} used network compression operations to compress well-trained networks. These methods allow transferring knowledge from previously trained networks and taking advantage of existing successful architectures in the target task, thus have shown improved efficiency and require significantly fewer computational resources (e.g., 5 GPUs in \citet{cai2018efficient}) to achieve competitive results. 

However, the network transformation operations in \citet{cai2018efficient} and \citet{ashok2017n2n} are still limited to only performing \emph{layer-level} architecture modifications such as adding (pruning) filters or inserting (removing) a layer, which does not change the topology of connection paths in a neural network. Hence, they restrict the search space to having the same path topology as the start network, i.e. they would always lead to chain-structured networks when given a chain-structured start point. As the state-of-the-art convolutional neural network (CNN) architectures have gone beyond simple chain-structured layout and demonstrated the effectiveness of multi-path structures such as Inception models \cite{szegedy2015going}, ResNets \cite{he2016deep} and DenseNets \cite{huang2016densely}, we would hope such methods to have the ability to explore a search space with different and complex path topologies while keeping the benefits of reusing weights.

% propose our work
In this paper, we present a new kind of transformation operations for neural networks, phrased as \emph{path-level} network transformation operations, which allows modifying the path topologies in a given network while allowing weight reusing to preserve the functionality like Net2Net operations \cite{chen2015net2net}. Based on the proposed path-level operations, we introduce a simple yet highly expressive tree-structured architecture space that can be viewed as a generalized version of multi-branch structures. To efficiently explore the introduced tree-structured architecture space, we further propose a bidirectional tree-structured \cite{tai2015improved} reinforcement learning meta-controller that can naturally encode the input tree, instead of simply using the chain-structured recurrent neural network \cite{zoph2017learning}. 

% TODO to be modified according to new exp results
% exp results 
Our experiments of learning CNN cells on CIFAR-10 show that our method using restricted computational resources (about 200 GPU-hours) can design highly effective cell structures. When combined with state-of-the-art human-designed architectures such as DenseNets \cite{huang2016densely} and PyramidNets \cite{han2016deep}, the best discovered cell shows significantly improved parameter efficiency and better results compared to the original ones. Specifically, without any additional regularization techniques, it achieves 3.14\% test error with 5.7M parameters, while DensNets give a best test error rate of 3.46\% with 25.6M parameters and PyramidNets give 3.31\% with 26.0M parameters. And with additional regularization techniques (DropPath \cite{zoph2017learning} and Cutout \cite{devries2017improved}), it reaches 2.30\% test error with 14.3M parameters, surpassing 2.40\% given by NASNet-A \cite{zoph2017learning} with 27.6M parameters and a similar training scheme. More importantly, NASNet-A is achieved using 48,000 GPU-hours while we only use 200 GPU-hours.
We further apply the best learned cells on CIFAR-10 to the ImageNet dataset by combining it with CondenseNet \cite{huang2017condensenet} for the $Mobile$ setting and also observe improved results when compared to models in the mobile setting.

\section{Related Work and Background}
% architecture search
\subsection{Architecture Search}\label{sec:arch_search_related}
Architecture search that aims to automatically find effective model architectures in a given architecture space has been studied using various approaches which can be categorized as neuro-evolution \cite{real2017large,liu2017hierarchical}, Bayesian optimization \cite{domhan2015speeding,mendoza2016towards}, Monte Carlo Tree Search \cite{negrinho2017deeparchitect} and reinforcement learning (RL) \cite{zoph2016neural,baker2016designing,zhong2017practical,zoph2017learning}.

Since getting an evaluation of each designed architecture requires training on the real data, which makes directly applying architecture search methods on large datasets (e.g., ImageNet \cite{deng2009imagenet}) computationally expensive, \citet{zoph2017learning} proposed to search for CNN cells that can be stacked later, rather than search for the entire architectures. Specifically, learning of the cell structures is conducted on small datasets (e.g., CIFAR-10) while learned cell structures are then transferred to large datasets (e.g., ImageNet). This scheme has also been incorporated in \citet{zhong2017practical} and \citet{liu2017hierarchical}. 

On the other hand, instead of constructing and evaluating architectures from scratch, there are some recent works that proposed to take network transformation operations to explore the architecture space given a trained network in the target task and reuse the weights. \citet{cai2018efficient} presented a recurrent neural network to iteratively generate transformation operations to be performed based on the current network architecture, and trained the recurrent network with REINFORCE algorithm \cite{williams1992simple}. A similar framework has also been incorporated in \citet{ashok2017n2n} where the transformation operations change from Net2Net operations in \citet{cai2018efficient} to compression operations. 

Compared to above work, in this paper, we extend current network transformation operations from layer-level to path-level. Similar to \citet{zoph2017learning} and \citet{zhong2017practical}, we focus on learning CNN cells, while our approach can be easily combined with any existing well-designed architectures to take advantage of their success and allow reusing weights to preserve the functionality. 
 
% multi-branch convolutional neural networks
\subsection{Multi-Branch Neural Networks}\label{sec:multi_branch}
Multi-branch structure (or motif) is an essential component in many modern state-of-the-art CNN architectures. The family of Inception models \cite{szegedy2015going,szegedy2017inception,szegedy2016rethinking} are successful multi-branch architectures with carefully customized branches.  
ResNets \cite{he2016deep} and DenseNets \cite{huang2016densely} can be viewed as two-branch architectures where one branch is the identity mapping. 
A common strategy within these multi-branch architectures is that the input feature map $\bx$ is first distributed to each branch based on a specific allocation scheme (either $split$ in Inception models or $replication$ in ResNets and DenseNets), then transformed by primitive operations (e.g., convolution, pooling, etc.) on each branch, and finally aggregated to produce an output based on a specific merge scheme (either $add$ in ResNets or $concatenation$ in Inception models and DenseNets).
%, as illustrated in Figure~TODO. % TODO An illustration of current multi-branch strutures here.

According to the research of \citet{veit2016residual}, ResNets can be considered to behave as ensembles of a collection of many paths of differing length. Similar interpretations can also be applied to Inception models and DenseNets. As the Inception models have demonstrated the merits of carefully customized branches where different primitive operations are used in each branch, it is thus of great interest to investigate whether we can benefit from more complex and well-designed path topologies within a CNN cell that make the collection of paths from the ensemble view more abundant and diverse.

In this work, we explore a tree-structured architecture space where at each node the input feature map is allocated to each branch, going through some primitive operations and the corresponding child node, and is later merged to produce an output for the node. It can be viewed as a generalization of current multi-branch architectures (tree with a depth of 1) and is able to embed plentiful paths within a CNN cell. 

\begin{figure}[t]
	\centering
	\includegraphics[width=0.96\columnwidth]{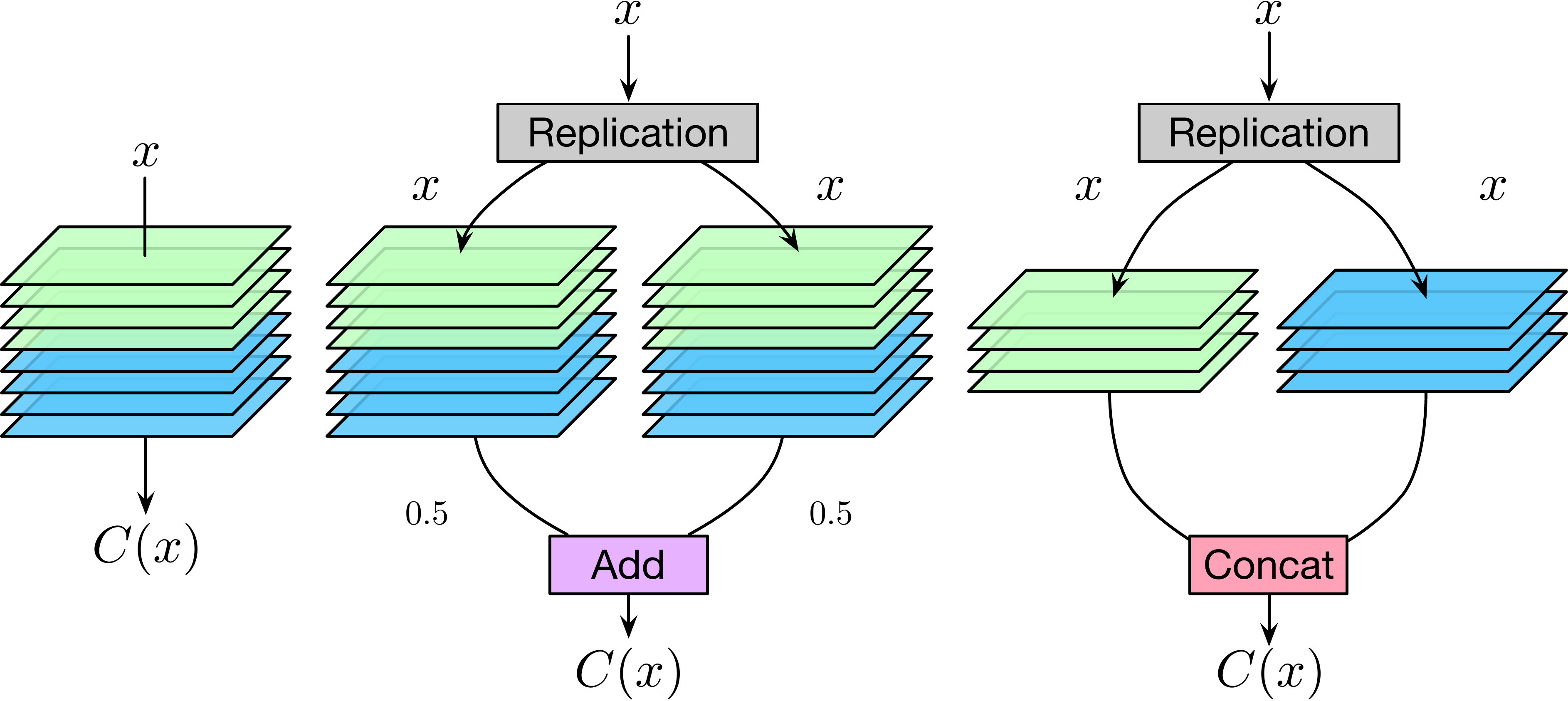}
	\caption{Convolution layer and its equivalent multi-branch motifs.}
	\label{fig:path_trans_conv}
\end{figure}

% function-preserving network transformation
\subsection{Function-Preserving Network Transformation}\label{para:net-trans}
Function-preserving network transformation refers to the class of network transformation operations that initialize a student network to preserve the functionality of a given teacher network. Net2Net technique \cite{chen2015net2net} introduces two specific function-preserving transformation operations, namely Net2WiderNet operation which replaces a layer with an equivalent layer that is wider (e.g., more filters for convolution layer) and Net2DeeperNet operation which replaces an identity mapping with a layer that can be initialized to be identity, including normal convolution layers with various filters (e.g., 3 $\times$ 3, 7 $\times$ 1, 1 $\times$ 7, etc.), depthwise-separable convolution layers \cite{chollet2016xception} and so on. Additionally, network compression operations \cite{han2015learning} that prune less important connections (e.g., low weight connections) to shrink the size of a given model without reducing the performance can also be viewed as one kind of function-preserving transformation operations. 

Our approach builds on existing function-preserving transformation operations and further extends to path-level architecture modifications. 

\section{Method}
\subsection{Path-Level Network Transformation}\label{sec:path_trans}
\begin{figure}[t]
	\centering
	\includegraphics[width=1\columnwidth]{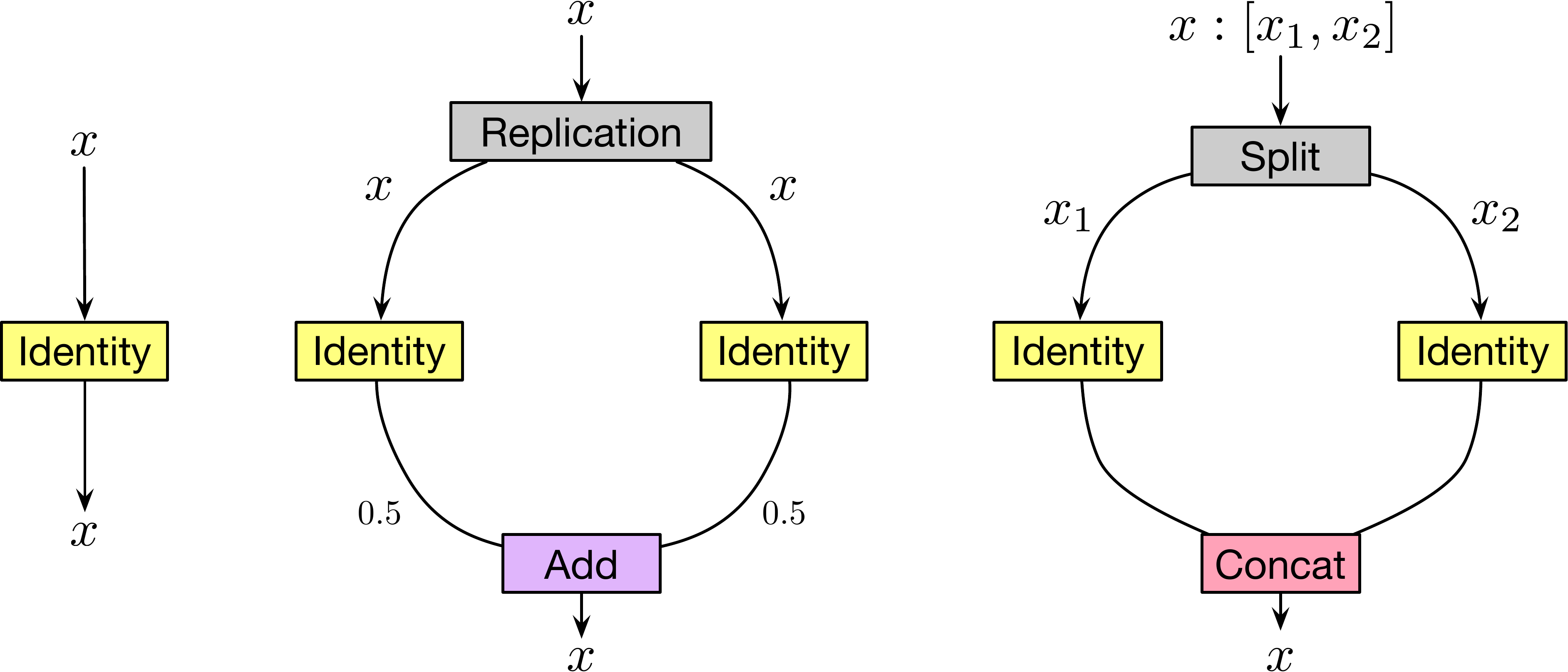}
	\caption{Identity layer and its equivalent multi-branch motifs.}
	\label{fig:path_trans_identity}
\end{figure}

% from single path to multi-branch
We introduce operations that allow replacing a single layer with a multi-branch motif whose merge scheme is either $add$ or $concatenation$. To illustrate the operations, we use two specific types of layers, i.e. identity layer and normal convolution layer, as examples, while they can also be applied to other similar types of layers, such as depthwise-separable convolution layers, analogously. 

For a convolution layer, denoted as $C(\cdot)$, to construct an equivalent multi-branch motif with $N$ branches, we need to set the branches so as to mimic the output of the original layer for any input feature map $\bx$. When these branches are merged by $add$, the allocation scheme is set to be $replication$ and we set each branch to be a replication of the original layer $C(\cdot)$, which makes each branch produce the same output (i.e. $C(\bx)$), and finally results in an output $N \times C(\bx)$ after being merged by $add$. To eliminate the factor, we further divide the output of each branch by $N$. As such the output of the multi-branch motif keeps the same as the output of the original convolution layer, as illustrated in Figure~\ref{fig:path_trans_conv} (middle).
When these branches are merged by $concatenation$, the allocation scheme is also set to be $replication$. Then we split the filters of the original convolution layer into $N$ parts along the output channel dimension and assign each part to the corresponding branch, which is later merged to produce an output $C(\bx)$, as shown in Figure~\ref{fig:path_trans_conv} (right).

\begin{figure*}[t]
	\centering
	\includegraphics[width=0.81\textwidth]{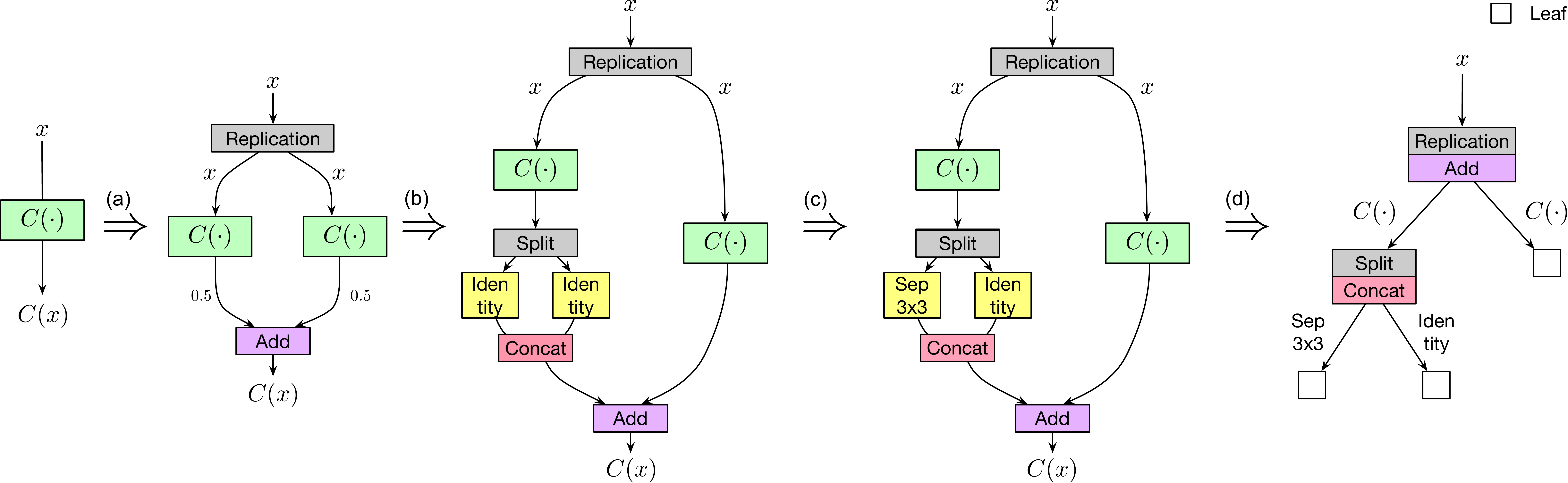}
    \vspace{-10pt}
	\caption{An illustration of transforming a single layer to a tree-structured motif via path-level transformation operations, where we apply Net2DeeperNet operation to replace an identity mapping with a $3 \times 3$ depthwise-separable convolution in (c). %\weinan{suggest to add $C(x)$ at the bottom of the middle three structures.}
	}\label{fig:to_tree}
\end{figure*}

For an identity layer, when the branches are merged by $add$, the transformation is the same except that the convolution layer in each branch changes to the identity mapping in this case (Figure~\ref{fig:path_trans_identity} (middle)). When the branches are merged by $concatenation$, the allocation scheme is set to be $split$ and each branch is set to be the identity mapping, as is illustrated in Figure~\ref{fig:path_trans_identity} (right).

Note that simply applying the above transformations does not lead to non-trivial path topology modifications. However, when combined with Net2Net operations, we are able to dramatically change the path topology, as shown in Figure~\ref{fig:to_tree}. For example, we can insert different numbers and types of layers into each branch by applying Net2DeeperNet operation, which makes each branch become substantially different, like Inception Models. Furthermore, since such transformations can be repetitively applied on any applicable layers in the neural network, such as a layer in the branch, we can thus arbitrarily increase the complexity of the path topology.

\subsection{Tree-Structured Architecture Space}\label{sec:tree_arch_space}

In this section, we describe the tree-structured architecture space that can be explored with path-level network transformation operations as illustrated in Figure~\ref{fig:to_tree}. 

A tree-structured architecture consists of edges and nodes, where at each node (except leaf nodes) we have a specific combination of the allocation scheme and the merge scheme, and the node is connected to each of its child nodes via an edge that is defined as a primitive operation such as convolution, pooling, etc.
Given the input feature map $\bx$, the output of node $N(\cdot)$, with $m$ child nodes denoted as $\{N^c_i(\cdot)\}$ and $m$ corresponding edges denoted as $\{E_i(\cdot)\}$, is defined recursively based on the outputs of its child nodes:
{\small
\begin{align}\label{eq:tree_architecture}
	\bz_i &= allocation(\bx, i), \nonumber \\
	\by_i &= N^c_i(E_i(\bz_i)), \quad 1 \leq i \leq m, \\
	N(x) &= merge(\by_1, \cdots, \by_m), \nonumber
\end{align}
}where $allocation(\bx, i)$ denotes the allocated feature map for $i^{th}$ child node based on the allocation scheme, and $merge(\cdot)$ denotes the merge scheme that takes the outputs of child nodes as input and outputs an aggregated result which is also the output of the node. For a leaf node that has no child node, it simply returns the input feature map as its output. As defined in Eq.~(\ref{eq:tree_architecture}), for a tree-structured architecture, the feature map is first fed to its root node, then spread to all subsequent nodes through allocation schemes at the nodes and edges in a top-down manner until reaching leaf nodes, and finally aggregated in mirror from the leaf nodes to the root node in a bottom-up manner to produce a final output feature map.
%\weinan{Question: whether the top-down allocation and bottom-up merge schemes should be in mirror?} \han{add "in mirror" after aggregated}

Notice that the tree-structured architecture space is not the full architecture space that can be achieved with the proposed path-level transformation operations. We choose to explore the tree-structure architecture space for the ease of implementation and further applying architecture search methods such as RL based approaches \cite{cai2018efficient} that would need to encode the architecture. Another reason for choosing the tree-structured architecture space is that it has a strong connection to existing multi-branch architectures, which can be viewed as tree-structured architectures with a depth of 1, i.e. all of the root node's child nodes are leaf. 

To apply architecture search methods on the tree-structured architecture space, we need to further specify it by defining the set of possible allocation schemes, merge schemes and primitive operations. As discussed in Sections~\ref{sec:multi_branch} and \ref{sec:path_trans}, the allocation scheme is either $replication$ or $split$ and the merge scheme is either $add$ or $concatenation$. For the primitive operations, similar to previous work \cite{zoph2017learning,liu2017hierarchical}, we consider the following 7 types of layers:
{\small
\begin{itemize}
	\setlength\itemsep{0.pt}
	\item $1 \times 1$ convolution
	\item Identity
	\item $3 \times 3$ depthwise-separable convolution 
	\item $5 \times 5$ depthwise-separable convolution
	\item $7 \times 7$ depthwise-separable convolution
    \item $3 \times 3$ average pooling
    \item $3 \times 3$ max pooling
\end{itemize}
}
Here, we include pooling layers that cannot be initialized as identity mapping. To preserve the functionality when pooling layers are chosen, we further reconstruct the weights in the student network (i.e. the network after transformations) to mimic the output logits of the given teacher network, using the idea of knowledge distillation \cite{hinton2015distilling}. As pooling layers do not dramatically destroy the functionality for multi-path neural networks, we find that the reconstruction process can be done with negligible cost. 

\subsection{Architecture Search with Path-Level Operations}\label{sec:arch_search_method}
In this section, we present an RL agent as the meta-controller to explore the tree-structured architecture space. The overall framework is similar to the one proposed in \citet{cai2018efficient} where the meta-controller iteratively samples network transformation actions to generate new architectures that are later trained to get the validation performances as reward signals to update the meta-controller via policy gradient algorithms.
To map the input architecture to transformation actions, the meta-controller has an encoder network that learns a low-dimensional representation of the given architecture, and distinct softmax classifiers that generate corresponding network transformation actions. 

%\subsubsection{Bidirectional Tree-structured Encoder}
In this work, as the input architecture now has a tree-structured topology that cannot be easily specified with a sequence of tokens, 
instead of using the chain-structure Long Short-Term Memory (LSTM) network \cite{hochreiter1997long} to encode the architecture \cite{zoph2017learning}, we propose to use a tree-structured LSTM. \citet{tai2015improved} introduced two kinds of tree-structured LSTM units, i.e. Child-Sum Tree-LSTM unit for tree structures whose child nodes are unordered and N-ary Tree-LSTM unit for tree structures whose child nodes are ordered. For further details, we refer to the original paper \cite{tai2015improved}.

\begin{figure}[t]
	\centering
	\begin{subfigure}{0.46\linewidth}
		\centering
		\includegraphics[height=1\columnwidth]{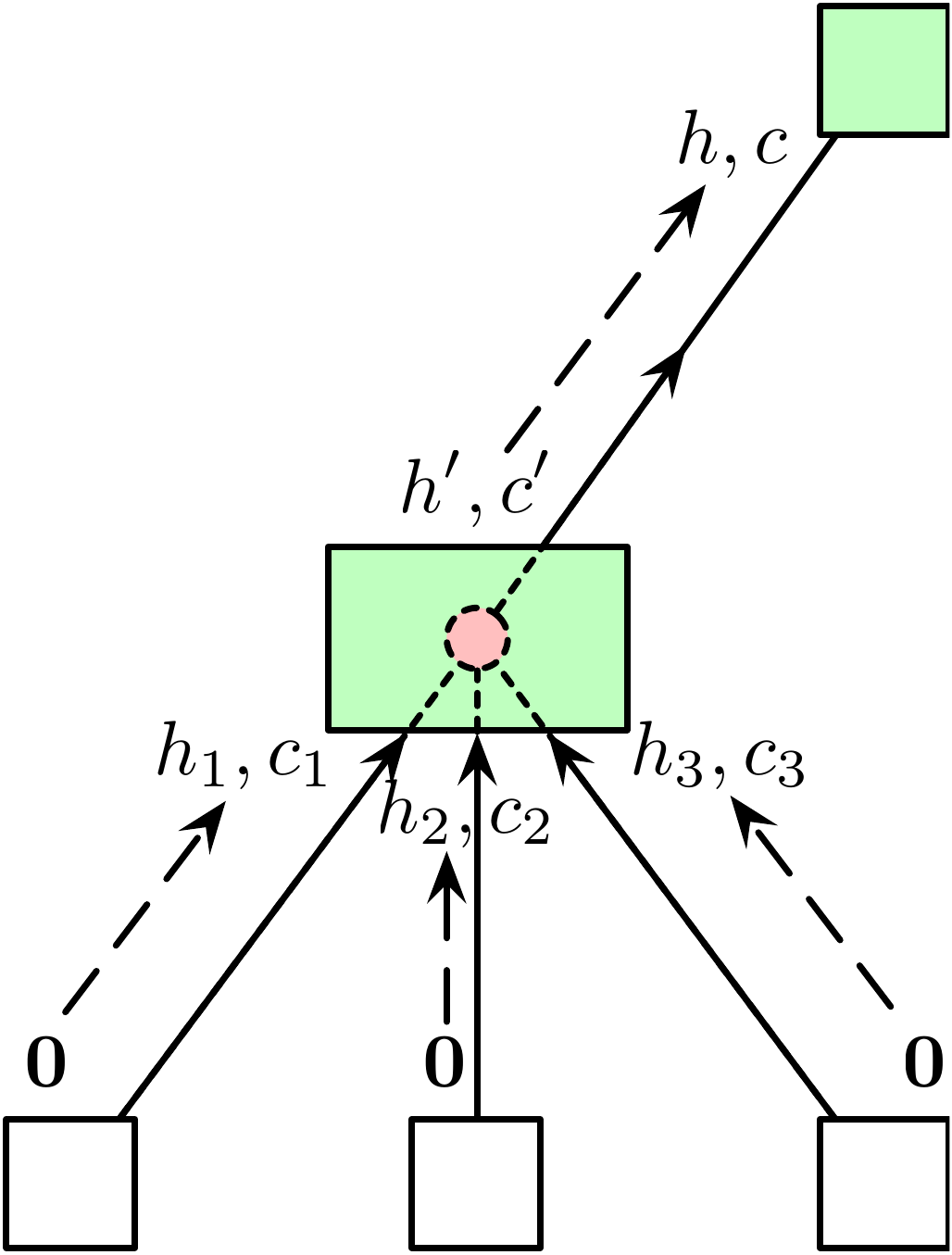}
		\vspace{-2pt}
		\caption{Bottom-up}
		\label{fig:encoder_bottom_up}
		\vspace{-0pt}
	\end{subfigure}
	\hfill
	\begin{subfigure}{0.46\linewidth}
		\centering
		\includegraphics[height=1\columnwidth]{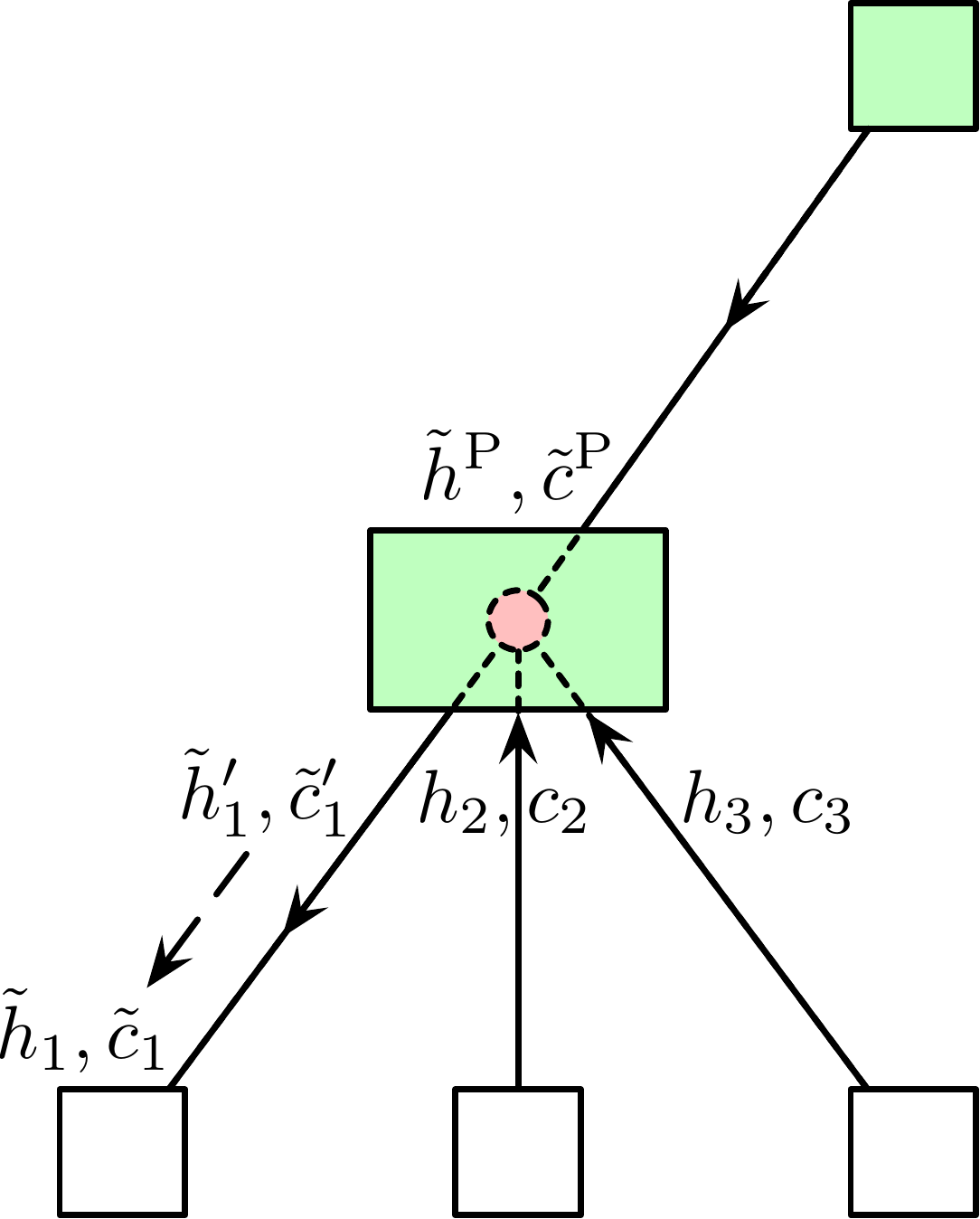}
		\vspace{-2pt}
		\caption{Top-down}
		\label{fig:encoder_top_down}
		\vspace{-0pt}
	\end{subfigure}	
	\caption{Calculation procedure of bottom-up and top-down hidden states.}
	\label{fig:tree_encoder}
\end{figure}

In our case, for the node whose merge scheme is $add$, its child nodes are unordered and thereby the Child-Sum Tree-LSTM unit is applied, while for the node whose merge scheme is $concatenation$, the N-ary Tree-LSTM unit is used since its child nodes are ordered. Additionally, as we have edges between nodes, we incorporate another normal LSTM unit for performing hidden state transitions on edges. We denote these three LSTM units as $ChildSumLSTM^\uparrow$, $NaryLSTM^\uparrow$ and $LSTM^\uparrow$, respectively. As such, the hidden state of the node that has $m$ child nodes is given as
{\small
\begin{align}\label{eq:bottom_up}
	h'\!\!, c' \!\! &= \!\! \begin{cases}
        \! ChildSumLSTM^\uparrow \! (s, [h_1^c, c_1^c], \cdots \!, [h_m^c, c_m^c]) & \!\!\!\! \text{if add} \\
        \! NaryLSTM^\uparrow \! (s, [h_1^c, c_1^c], \cdots \!, [h_m^c, c_m^c]) & \!\!\!\! \text{if concat}
    \end{cases} \! , \nonumber \\
	h, c &= LSTM^\uparrow(e, [h', c']) ,
\end{align}
}where $[h^c_i, c^c_i]$ denotes the hidden state
of $i^{th}$ child node, $s$ represents the allocation and merge scheme of the node, $e$ is the edge that connects the node to its parent node, and $[h, c]$ is the hidden state of the node. Such calculation is done in a bottom-up manner as is shown in Figure~\ref{fig:encoder_bottom_up}.

Note that the hidden state calculated via Eq.~(\ref{eq:bottom_up}) only contains information below the node. Analogous to bidirectional LSTM, we further consider a top-down procedure, using two new LSTM units ($NaryLSTM^\downarrow$ and $LSTM^\downarrow$), to calculate another hidden state for each node. We refer to these two hidden states of a node as bottom-up hidden state and top-down hidden state respectively. For a node, with $m$ child nodes, whose top-down hidden state is $[\tilde{h}^p, \tilde{c}^p]$, the top-down hidden state of its $i^{th}$ child node is given as
{\small
\begin{align}\label{eq:top_down}
	\tilde{h}'_i, \tilde{c}'_i &= NaryLSTM^\downarrow(s, [\tilde{h}^p, \tilde{c}^p], [h_1, c_1], \cdots,  \!\!\!\! \overbrace{[\bs{0}, \bs{0}]}^{\text{$i^{th}$ child node}} \!\!\!\!, \cdots), \nonumber \\
	\tilde{h}_i, \tilde{c}_i &= LSTM^\downarrow(e_i, [\tilde{h}'_i, \tilde{c}'_i]),
\end{align}
}where $[h_j, c_j]$ is the bottom-up hidden state of $j^{th}$ child node, $s$ is the allocation and merge scheme of the node, $e_i$ is the edge that connect the node to its $i^{th}$ child node, and $[\tilde{h}_i, \tilde{c}_i]$ is the top-down hidden state of $i^{th}$ child node. As shown in Figure~\ref{fig:encoder_top_down} and Eq.~(\ref{eq:top_down}), a combination of the bottom-up hidden state and top-down hidden state now forms a comprehensive hidden state for each node, containing all information of the architecture. 

\begin{figure}[t]
	\centering
	\begin{subfigure}{0.32\linewidth}
		\centering
		\includegraphics[width=1.1\columnwidth]{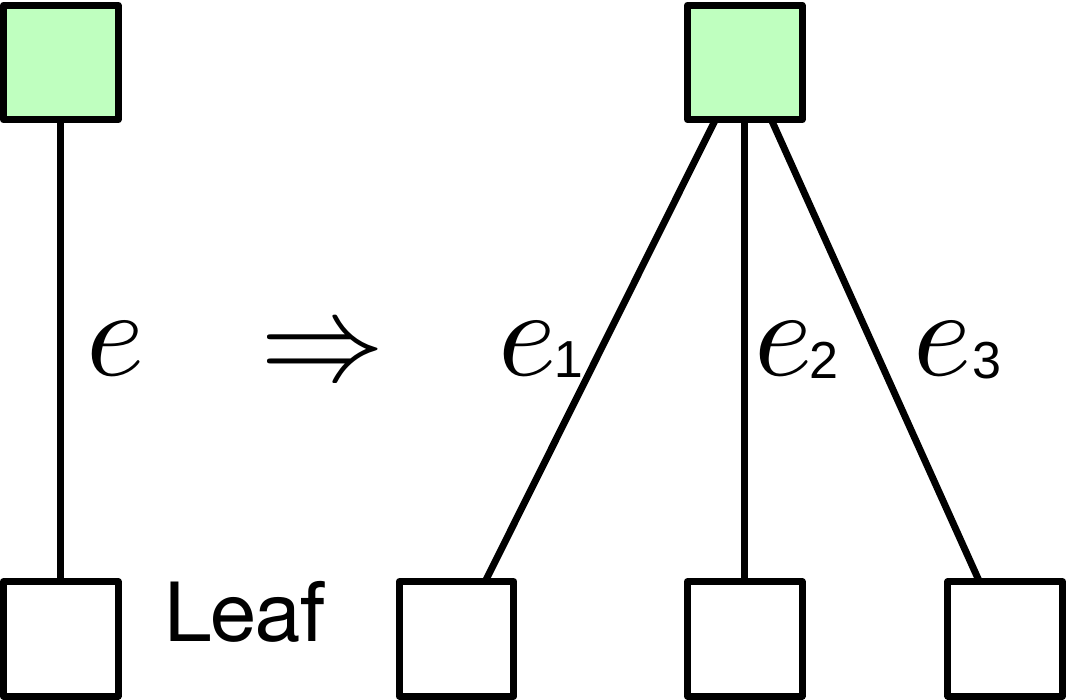}
		\caption{}
		\label{fig:action-a}
		\vspace{-0pt}
	\end{subfigure}
	%\hfill
	\begin{subfigure}{0.32\linewidth}
		\centering
		\hspace{10pt}
		\includegraphics[width=1.1\columnwidth]{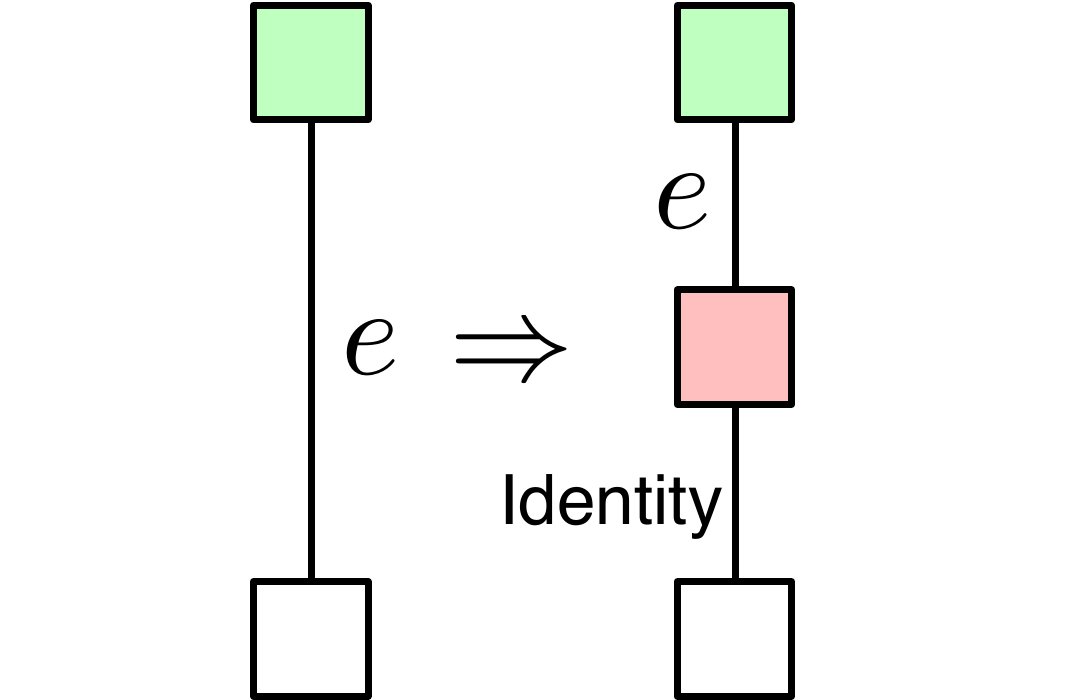}
		\caption{}
		\label{fig:action-b}
		\vspace{-0pt}
	\end{subfigure}	
	%\hfill
	\begin{subfigure}{0.32\linewidth}
		\centering
		\includegraphics[width=1.1\columnwidth]{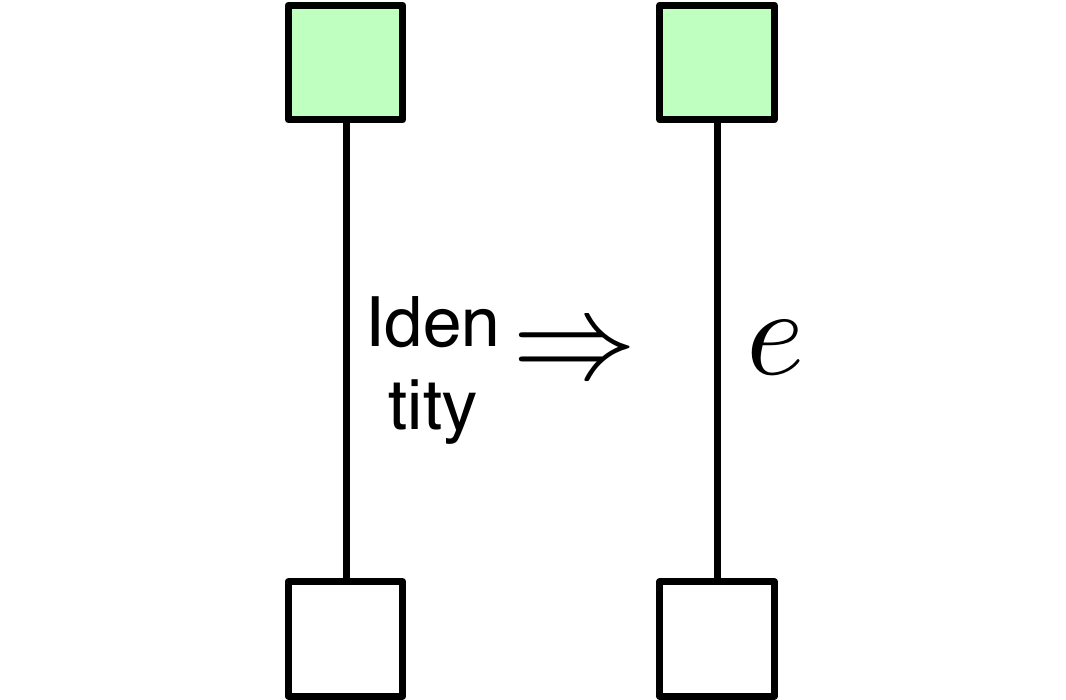}
		\caption{}
		\label{fig:action-c}
		\vspace{-0pt}
	\end{subfigure}	
	\caption{
		Illustration of transformation decisions on nodes and edges. (a) The meta-controller transforms a node with only one leaf child node to have multiple child nodes. Both merge scheme and branch number are predicted.  (b) The meta-controller inserts a new leaf node to be the child node of a previous leaf node and they are connected with an identity mapping. (c) The meta-controller replaces an identity mapping with a layer (can be identity) chosen from the set of possible primitive operations.
	}
	\label{fig:actions}
\end{figure}

Given the hidden state at each node, we have various softmax classifiers for making different transformation decisions on applicable nodes as follows:

\begin{enumerate}
	\item For a node that has only one leaf child node, the meta-controller chooses a merge scheme from $\{add, concatenation, none\}$. When $add$ or $concatenation$ is chosen, the meta-controller further chooses the number of branches and then the network is transformed accordingly, which makes the node have multiple child nodes now (Figure~\ref{fig:action-a}). When $none$ is chosen, nothing is done and the meta-controller will not make such decision on that node again. 
	\item For a node that is a leaf node, the meta-controller determines whether to expand the node, i.e. insert a new leaf node to be the child node of this node and connect them with identity mapping, which increases the depth of the architecture (Figure~\ref{fig:action-b}).
	\item For an identity edge, the meta-controller chooses a new edge (can be identity) from the set of possible primitive operations (Section~\ref{sec:tree_arch_space}) to replace the identity edge (Figure~\ref{fig:action-c}). Also this decision will only be made once for each edge. 
\end{enumerate}

\begin{figure}[t]
	\centering
	\includegraphics[width=1\columnwidth]{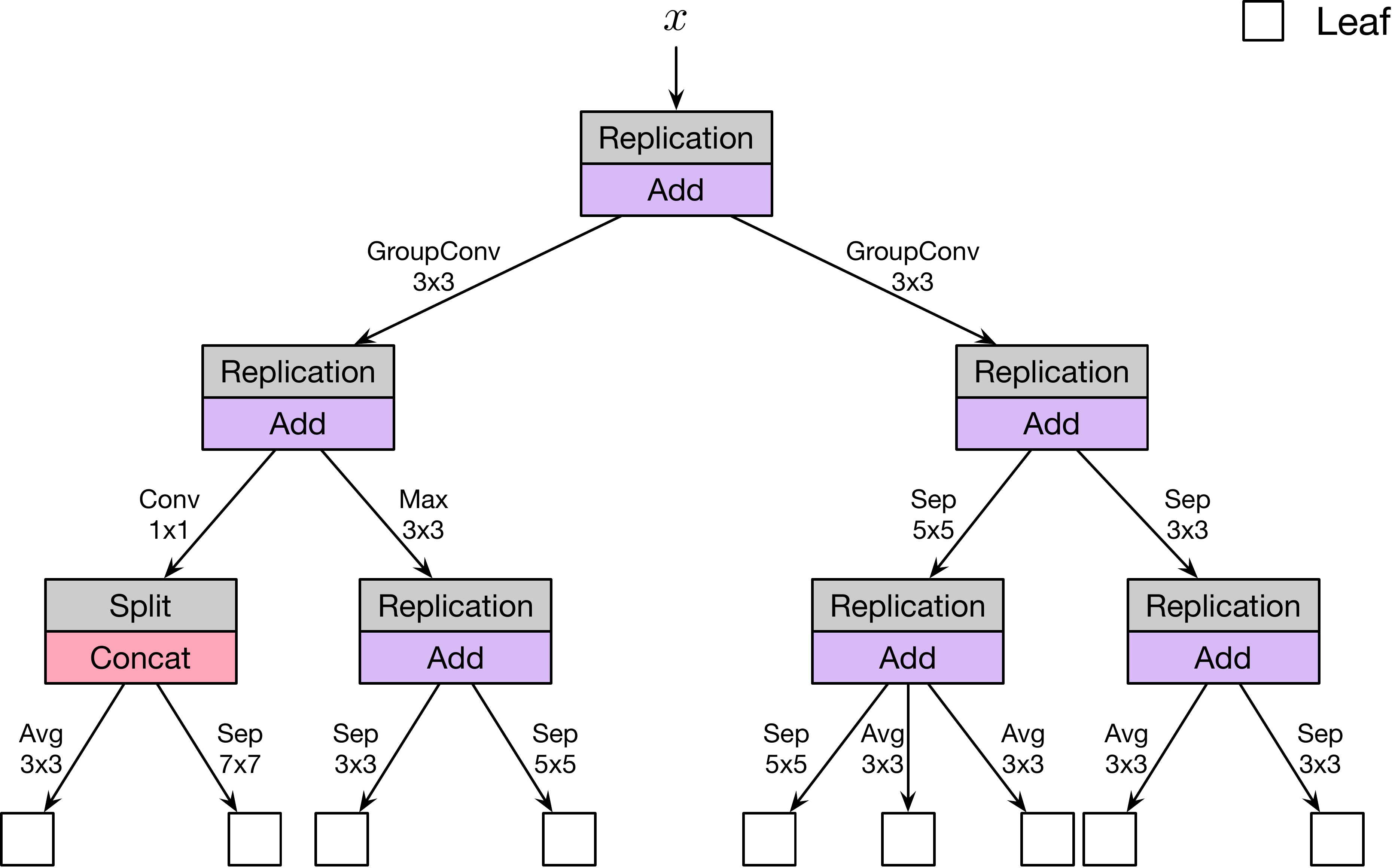}
    \vspace{-10pt}
	\caption{Detailed structure of the best discovered cell on CIFAR-10 (TreeCell-A). ``GroupConv'' denotes the group convolution; ``Conv'' denotes the normal convolution; ``Sep'' denotes the depthwise-separable convolution; ``Max'' denotes the max pooling; ``Avg'' denotes the average pooling.}
	\label{fig:tree-cell-a}
\end{figure}

\section{Experiments and Results}\label{sec:exp}
Our experimental setting\footnote{Experiment code: https://github.com/han-cai/PathLevel-EAS} resembles \citet{zoph2017learning}, \citet{zhong2017practical} and \citet{liu2017hierarchical}. Specifically, we apply the proposed method described above to learn CNN cells on CIFAR-10 \cite{krizhevsky2009learning} for the image classification task and transfer the learned cell structures to ImageNet dataset \cite{deng2009imagenet}.

\subsection{Experimental Details}
% dataset
CIFAR-10 contains 50,000 training images and 10,000 test images, where we randomly sample 5,000 images from the training set to form a validation set for the architecture search process, similar to previous work \cite{zoph2017learning,cai2018efficient}. We use a standard data augmentation scheme (mirroring/shifting) that is widely used for this dataset \cite{huang2016densely,han2016deep,cai2018efficient} and normalize the images using channel means and standard deviations for preprocessing.

% encoder network
For the meta-controller, described in Section~\ref{sec:arch_search_method}, the hidden state size of all LSTM units is 100 and we train it with the ADAM optimizer \cite{kingma2014adam} using the REINFORCE algorithm \cite{williams1992simple}. To reduce variance, we adopt a baseline function which is an exponential moving average of previous rewards with a decay of 0.95, as done in \citet{cai2018efficient}. We also use an entropy penalty with a weight of 0.01 to ensure exploration. 

\begin{figure}[t]
	\centering
	\includegraphics[width=\columnwidth]{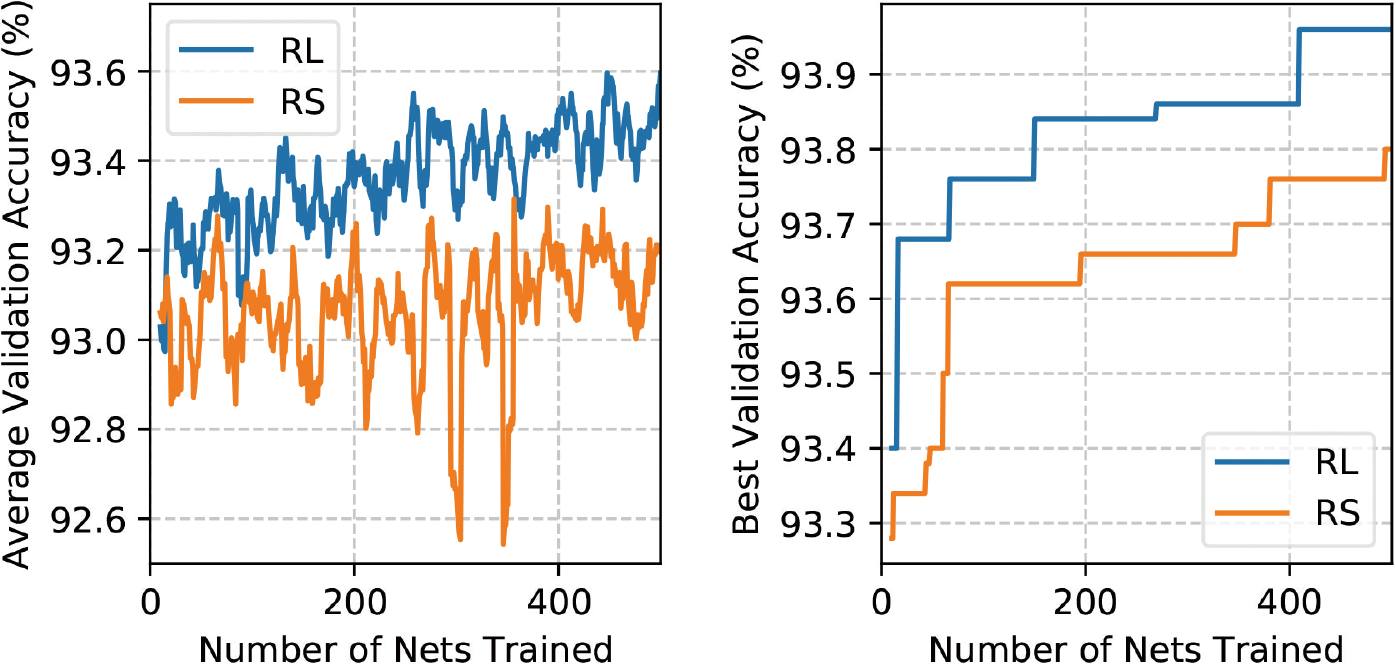}
    \vspace{-10pt}
	\caption{Progress of the architecture search process and comparison between RL and random search (RS) on CIFAR-10.}\vspace{-15pt}
	\label{fig:arch_search_progress}
\end{figure}

% training details
At each step in the architecture search process, the meta-controller samples a tree-structured cell by taking transformation actions starting with a single layer in the base network. For example, when using a DenseNet as the base network, after the transformations, all $3 \times 3$ convolution layers in the dense blocks are replaced with the sampled tree-structured cell while all the others remain unchanged. The obtained network, along with weights transferred from the base network, is then trained for 20 epochs on CIFAR-10 with an initial learning rate of 0.035 that is further annealed with a cosine learning rate decay \cite{loshchilov2016sgdr}, a batch size of 64, a weight decay of 0.0001, using the SGD optimizer with a Nesterov momentum of 0.9. The validation accuracy $acc_v$ of the obtained network is used to compute a reward signal. We follow \citet{cai2018efficient} and use the transformed value, i.e. $tan(acc_v \times \pi / 2)$, as the reward since improving the accuracy from 90\% to 91\% should gain much more than from 60\% to 61\%. Additionally, we update the meta-controller with mini-batches of 10 architectures. 
 
After the architecture search process is done, the learned cell structures can be embedded into various kinds of base networks (e.g., ResNets, DenseNets, etc.) with different depth and width. In this stage, we train networks for 300 epochs with an initial learning rate of 0.1, while all other settings keep the same. 

\subsection{Architecture Search on CIFAR-10}

\begin{table*}[t]
	\centering
	\caption{Test error rate (\%) results of our best discovered architectures as well as state-of-the-art human-designed and automatically designed architectures on CIFAR-10. If ``Reg'' is checked, additional regularization techniques (e.g., Shake-Shake \cite{gastaldi2017shake}, DropPath \cite{zoph2017learning} and Cutout \cite{devries2017improved}), along with a longer training schedule (600 epochs or 1800 epochs) are utilized when training the networks. 
	}\label{tab:res_on_cifar}
	\vspace{5pt}
	\resizebox{2\columnwidth}{!}{  
		\begin{tabular}{l | l | c | c | c }
			\hline
			& Model & Reg & Params & Test error \\
			\hline
			\tabincell{l}{Human \\ designed} & \tabincell{l}{ResNeXt-29 ($16 \times 64$d) \cite{xie2017aggregated} \\ DenseNet-BC ($N=31, k=40$) \cite{huang2016densely} \\ PyramidNet-Bottleneck ($N=18, \alpha = 270$) \cite{han2016deep} \\ PyramidNet-Bottleneck ($N=30, \alpha = 200$) \cite{han2016deep} \\ ResNeXt + Shake-Shake (1800 epochs) \cite{gastaldi2017shake} \\ ResNeXt + Shake-Shake + Cutout (1800 epochs) \cite{devries2017improved}} & \tabincell{c}{ \\ \\  \\ \\ $\checkmark$ \\ $\checkmark$} & \tabincell{c}{68.1M \\ 25.6M \\ 27.0M \\ 26.0M \\ 26.2M \\ 26.2M} & \tabincell{c}{3.58 \\ 3.46 \\ 3.48 \\ 3.31 \\ 2.86 \\ 2.56} \\
			\hline
			\tabincell{l}{Auto \\ designed} & \tabincell{l}{
            EAS (plain CNN) \cite{cai2018efficient} \\ Hierarchical ($c_0 = 128$) \cite{liu2017hierarchical} \\
            Block-QNN-A ($N = 4$) \cite{zhong2017practical} \\ NAS v3 \cite{zoph2016neural} \\ NASNet-A (6, 32) + DropPath (600 epochs) \cite{zoph2017learning} \\ NASNet-A (6, 32) + DropPath + Cutout (600 epochs) \cite{zoph2017learning} \\ NASNet-A (7, 96) + DropPath + Cutout (600 epochs) \cite{zoph2017learning}} & \tabincell{c}{ \\ \\ \\ \\ $\checkmark$ \\ $\checkmark$ \\ $\checkmark$} & \tabincell{c}{23.4M \\ - \\ - \\ 37.4M \\ 3.3M \\ 3.3M \\ 27.6M} & \tabincell{c}{4.23 \\ 3.63 \\ 3.60 \\ 3.65 \\ 3.41 \\ 2.65 \\ 2.40} \\
			\hline
			\tabincell{l}{Ours} & \tabincell{l}{TreeCell-B with DenseNet ($N=6, k=48, G=2$) \\ TreeCell-A with DenseNet ($N=6, k=48, G=2$) \\ TreeCell-A with DenseNet ($N=16, k=48, G=2$) \\ TreeCell-B with PyramidNet ($N=18, \alpha=84, G=2$) \\ TreeCell-A with PyramidNet ($N=18, \alpha=84, G=2$) \\ TreeCell-A with PyramidNet ($N=18, \alpha=84, G=2$) + DropPath (600 epochs) \\ TreeCell-A with PyramidNet ($N=18, \alpha=84, G=2$) + DropPath + Cutout (600 epochs) \\ TreeCell-A with PyramidNet ($N=18, \alpha=150, G=2$) + DropPath + Cutout (600 epochs)} & \tabincell{c}{ \\ \\ \\ \\ \\ $\checkmark$ \\ $\checkmark$ \\ $\checkmark$} & \tabincell{c}{ 3.2M \\ 3.2M \\ 13.1M \\ 5.6M \\ 5.7M \\ 5.7M \\ 5.7M \\ 14.3M} & \tabincell{c}{ 3.71 \\ 3.64 \\ 3.35 \\ 3.40 \\ 3.14 \\ 2.99 \\ 2.49 \\ \textbf{2.30}} \\
			\hline
		\end{tabular}
	}
\end{table*}

In our experiments, we use a small DenseNet-BC ($N = 2, L = 16, k = 48, G = 4$)\footnote{$N$, $L$ and $k$ respectively indicate the number of $3 \times 3$ convolution layers within each dense block, the depth of the network, and the growth rate, i.e. the number of filters of each $3 \times 3$ convolution layer. And we use the group convolution with $G = 4$ groups here.
For DenseNet-BC, $L = 6 \times N + 4$, so we omit $L$ in the following discussions for simplicity.}, 
which achieves an accuracy of $93.12\%$ on the held-out validation set, as the base network to learn cell structures. We set the maximum depth of the cell structures to be 3, i.e. the length of the path from the root node to each leaf node is no larger than 3 (Figure~\ref{fig:tree-cell-a}). For nodes whose merge scheme is $add$, the number of branches is chosen from $\{2, 3\}$ while for nodes whose merge scheme is $concatenation$, the number of branches is set to be 2. Additionally, we use very restricted computational resources for this experiment (about 200 GPU-hours $\ll$ 48,000 GPU-hours in \citet{zoph2017learning}) with in total 500 networks trained. 

The progress of the architecture search process is reported in Figure~\ref{fig:arch_search_progress}, where the results of random search (a very strong baseline for black-box optimization \cite{bergstra2012random}) under the same condition is also provided. We can find that the average validation accuracy of the designed architectures by the RL meta-controller gradually increases as the number of sampled architectures increases, as expected, while the curve of random search keeps fluctuating, which indicates that the RL meta-controller effectively focuses on the right search direction while random search fails. Therefore, with only 500 networks trained, the best model identified by RL, after 20 epochs training, achieves 0.16$\%$ better validation accuracy than the best model identified by random search. 

We take top 10 candidate cells discovered in this experiment, and embed them into a relatively larger base network, i.e. DenseNet-BC ($N=6, k=48, G$) where $G$ is chosen from $\{1, 2, 4\}$ to make different cells have a similar number of parameters as the normal $3 \times 3$ convolution layer (more details in the supplementary material). After 300 epochs training on CIFAR-10, the top 2 cells achieve 3.64\% test error (TreeCell-A) and 3.71\% test error (TreeCell-B), respectively. The detailed structure of TreeCell-A is given in Figure~\ref{fig:tree-cell-a}, while TreeCell-B's detailed structure is provided in the supplementary material. Under the same condition, the best cell given by random search reaches a test error rate of 3.98\%, which is 0.34\% worse than TreeCell-A.

%In addition to starting from DenseNet-BC ($N = 2, k = 48, G = 4$), we also conduct architecture search on CIFAR-10 starting from scratch (i.e. a chain of identity mappings) to learn how much we can benefit from reusing well-designed networks (more details in the supplementary material). The progress of this architecture search process is illustrated in Figure~\ref{fig:arch_search_progress_identity}, where we can observe similar trends 

\subsection{Results on CIFAR-10}
We further embed the top discovered cells, i.e. TreeCell-A and TreeCell-B, into larger base networks. Beside DenseNets, to justify whether the discovered cells starting with DenseNet can be transferred to other types of architectures such as ResNets, we also embed the cells into PyramidNets \cite{han2016deep}, a variant of ResNets. 

The summarized results are reported in Table~\ref{tab:res_on_cifar}. When combined with DenseNets, the best discovered tree cell (i.e. TreeCell-A) achieves a test error rate of 3.64\% with only 3.2M parameters, which is comparable to the best result, i.e. 3.46\% in the original DenseNet paper, given by a much larger DenseNet-BC with 25.6M parameters. Furthermore, the test error rate drops to 3.35\% as the number of parameters increases to 13.1M. We attribute the improved parameter efficiency and better test error rate results to the improved representation power from the increased path topology complexity introduced by the learned tree cells. 
When combined with PyramidNets, TreeCell-A reaches 3.14\% test error with only 5.7M parameters while the best PyramidNet achieves 3.31\% test error with 26.0M parameters, which also indicates significantly improved parameter efficiency by incorporating the learned tree cells for PyramidNets. Since the cells are learned using a DenseNet as the start point rather than a PyramidNet, it thereby justifies the transferability of the learned cells to other types of architectures. 

We notice that there are some strong regularization techniques that have shown to effectively improve the performances on CIFAR-10, such as Shake-Shake \cite{gastaldi2017shake}, DropPath \cite{zoph2017learning} and Cutout \cite{devries2017improved}. In our experiments, when using DropPath that stochastically drops out each path (i.e. edge in the tree cell) and training the network for 600 epochs, as done in \citet{zoph2017learning} and \citet{liu2017progressive}, TreeCell-A reaches 2.99\% test error with 5.7M parameters. Moreover, with Cutout, TreeCell-A further achieves 2.49\% test error with 5.7M parameters and 2.30\% test error with 14.3M parameters, outperforming all compared human-designed and automatically designed architectures on CIFAR-10 while having much fewer parameters (Table~\ref{tab:res_on_cifar}). 

We would like to emphasize that these results are achieved with only 500 networks trained using about 200 GPU-hours while the compared architecture search methods utilize much more computational resources to achieve their best results, such as \citet{zoph2017learning} that used 48,000 GPU-hours. 

\subsection{Results on ImageNet}
Following \citet{zoph2017learning} and \citet{zhong2017practical}, we further test the best cell structures learned on CIFAR-10, i.e. TreeCell-A and TreeCell-B, on ImageNet dataset. Due to resource and time constraints, we focus on the $Mobile$ setting in our experiments, where the input image size is $224 \times 224$ and we train relatively small models that require less than 600M multiply-add operations to perform inference on a single image. To do so, we combine the learned cell structures with CondenseNet \cite{huang2017condensenet}, a recently proposed efficient network architecture that is designed for the $Mobile$ setting. 
\begin{table}[t]
	\centering
	\caption{Top-1 (\%) and Top-5 (\%) classification error rate results on ImageNet in the $Mobile$ Setting ($\leq$ 600M multiply-add operations). ``$\times +$'' denotes the number of multiply-add operations.}\label{tab:res_on_imagenet}
	\vspace{5pt}
	\resizebox{\columnwidth}{!}{  
		\begin{tabular}{l | c | c | c }
			\hline
			Model & $\times +$ & Top-1 & Top-5 \\
			\hline
			1.0 MobileNet-224 \cite{howard2017mobilenets} & 569M & 29.4 & 10.5 \\
			ShuffleNet 2x \cite{zhang2017shufflenet} & 524M & 29.1 & 10.2 \\
			CondenseNet ($G_1=G_3=8$) \cite{huang2017condensenet} & 274M & 29.0 & 10.0 \\
			CondenseNet ($G_1=G_3=4$) \cite{huang2017condensenet} & 529M & 26.2 & 8.3 \\
			\hline
			NASNet-A ($N=4$) \cite{zoph2017learning} & 564M & 26.0 & 8.4 \\
			NASNet-B ($N=4$) \cite{zoph2017learning} & 448M & 27.2 & 8.7 \\
			NASNet-C ($N=3$) \cite{zoph2017learning} & 558M & 27.5 & 9.0 \\
			\hline
			TreeCell-A with CondenseNet ($G_1=4, G_3=8$) & 588M & 25.5 & 8.0 \\
            TreeCell-B with CondenseNet ($G_1=4, G_3=8$) & 594M & 25.4 & 8.1 \\
			\hline
		\end{tabular}
	}
\end{table}

The result is reported in Table~\ref{tab:res_on_imagenet}. By embedding TreeCell-A into CondenseNet ($G_1=4, G_3=8$) where each block comprises a learned $1 \times 1$ group convolution layer with $G_1 = 4$ groups and a standard $3 \times 3$ group convolution layer with $G_3 = 8$ groups, we achieve 25.5\% top-1 error and 8.0\% top-5 error with 588M multiply-add operations, which significantly outperforms MobileNet and ShuffleNet, and is also better than CondenseNet ($G_1=G_3=4$) with a similar number of multiply-add operations. Meanwhile, we find that TreeCell-B with CondenseNet ($G_1=4, G_3=8$) reaches a slightly better top-1 error result, i.e. 25.4\%, than TreeCell-A. 

When compared to NASNet-A, we also achieve slightly better results with similar multiply-add operations despite the fact that they used 48,000 GPU-hours to achieve these results while we only use 200 GPU-hours. By taking advantage of existing successful human-designed architectures, we can easily achieve similar (or even better) results with much fewer computational resources, compared to exploring the architecture space from scratch. 

\section{Conclusion}
In this work, we presented path-level network transformation operations as an extension to current function-preserving network transformation operations to enable the architecture search methods to perform not only layer-level architecture modifications but also path-level topology modifications in a neural network. Based on the proposed path-level transformation operations, we further explored a tree-structured architecture space, a generalized version of current multi-branch architectures, that can embed plentiful paths within each CNN cell, with a bidirectional tree-structured RL meta-controller. The best designed cell structure by our method using only 200 GPU-hours has shown both improved parameter efficiency and better test accuracy on CIFAR-10, when combined with state-of-the-art human designed architectures including DenseNets and PyramidNets. And it has also demonstrated its transferability on ImageNet dataset in the $Mobile$ setting. For future work, we would like to combine the proposed method with network compression operations to explore the architecture space with the model size and the number of multiply-add operations taken into consideration and conduct experiments on other tasks such as object detection.

\bibliography{path-trans}

\begin{thebibliography}{41}
\providecommand{\natexlab}[1]{#1}
\providecommand{\url}[1]{\texttt{#1}}
\expandafter\ifx\csname urlstyle\endcsname\relax
  \providecommand{\doi}[1]{doi: #1}\else
  \providecommand{\doi}{doi: \begingroup \urlstyle{rm}\Url}\fi

\bibitem[Ashok et~al.(2018)Ashok, Rhinehart, Beainy, and Kitani]{ashok2017n2n}
Ashok, A., Rhinehart, N., Beainy, F., and Kitani, K.~M.
\newblock N2n learning: Network to network compression via policy gradient
  reinforcement learning.
\newblock \emph{ICLR}, 2018.

\bibitem[Baker et~al.(2017)Baker, Gupta, Naik, and Raskar]{baker2016designing}
Baker, B., Gupta, O., Naik, N., and Raskar, R.
\newblock Designing neural network architectures using reinforcement learning.
\newblock \emph{ICLR}, 2017.

\bibitem[Bergstra \& Bengio(2012)Bergstra and Bengio]{bergstra2012random}
Bergstra, J. and Bengio, Y.
\newblock Random search for hyper-parameter optimization.
\newblock \emph{Journal of Machine Learning Research}, 2012.

\bibitem[Cai et~al.(2018)Cai, Chen, Zhang, Yu, and Wang]{cai2018efficient}
Cai, H., Chen, T., Zhang, W., Yu, Y., and Wang, J.
\newblock Efficient architecture search by network transformation.
\newblock In \emph{AAAI}, 2018.

\bibitem[Chen et~al.(2016)Chen, Goodfellow, and Shlens]{chen2015net2net}
Chen, T., Goodfellow, I., and Shlens, J.
\newblock Net2net: Accelerating learning via knowledge transfer.
\newblock \emph{ICLR}, 2016.

\bibitem[Chollet(2016)]{chollet2016xception}
Chollet, F.
\newblock Xception: Deep learning with depthwise separable convolutions.
\newblock \emph{arXiv preprint}, 2016.

\bibitem[Deng et~al.(2009)Deng, Dong, Socher, Li, Li, and
  Fei-Fei]{deng2009imagenet}
Deng, J., Dong, W., Socher, R., Li, L.-J., Li, K., and Fei-Fei, L.
\newblock Imagenet: A large-scale hierarchical image database.
\newblock In \emph{CVPR}, 2009.

\bibitem[DeVries \& Taylor(2017)DeVries and Taylor]{devries2017improved}
DeVries, T. and Taylor, G.~W.
\newblock Improved regularization of convolutional neural networks with cutout.
\newblock \emph{arXiv preprint arXiv:1708.04552}, 2017.

\bibitem[Domhan et~al.(2015)Domhan, Springenberg, and
  Hutter]{domhan2015speeding}
Domhan, T., Springenberg, J.~T., and Hutter, F.
\newblock Speeding up automatic hyperparameter optimization of deep neural
  networks by extrapolation of learning curves.
\newblock In \emph{IJCAI}, 2015.

\bibitem[Gastaldi(2017)]{gastaldi2017shake}
Gastaldi, X.
\newblock Shake-shake regularization.
\newblock \emph{arXiv preprint arXiv:1705.07485}, 2017.

\bibitem[Han et~al.(2017)Han, Kim, and Kim]{han2016deep}
Han, D., Kim, J., and Kim, J.
\newblock Deep pyramidal residual networks.
\newblock \emph{CVPR}, 2017.

\bibitem[Han et~al.(2015)Han, Pool, Tran, and Dally]{han2015learning}
Han, S., Pool, J., Tran, J., and Dally, W.
\newblock Learning both weights and connections for efficient neural network.
\newblock In \emph{NIPS}, 2015.

\bibitem[He et~al.(2016)He, Zhang, Ren, and Sun]{he2016deep}
He, K., Zhang, X., Ren, S., and Sun, J.
\newblock Deep residual learning for image recognition.
\newblock In \emph{CVPR}, 2016.

\bibitem[Hinton et~al.(2015)Hinton, Vinyals, and Dean]{hinton2015distilling}
Hinton, G., Vinyals, O., and Dean, J.
\newblock Distilling the knowledge in a neural network.
\newblock \emph{arXiv preprint arXiv:1503.02531}, 2015.

\bibitem[Hochreiter \& Schmidhuber(1997)Hochreiter and
  Schmidhuber]{hochreiter1997long}
Hochreiter, S. and Schmidhuber, J.
\newblock Long short-term memory.
\newblock \emph{Neural computation}, 1997.

\bibitem[Howard et~al.(2017)Howard, Zhu, Chen, Kalenichenko, Wang, Weyand,
  Andreetto, and Adam]{howard2017mobilenets}
Howard, A.~G., Zhu, M., Chen, B., Kalenichenko, D., Wang, W., Weyand, T.,
  Andreetto, M., and Adam, H.
\newblock Mobilenets: Efficient convolutional neural networks for mobile vision
  applications.
\newblock \emph{arXiv preprint arXiv:1704.04861}, 2017.

\bibitem[Huang et~al.(2017{\natexlab{a}})Huang, Liu, van~der Maaten, and
  Weinberger]{huang2017condensenet}
Huang, G., Liu, S., van~der Maaten, L., and Weinberger, K.~Q.
\newblock Condensenet: An efficient densenet using learned group convolutions.
\newblock \emph{arXiv preprint arXiv:1711.09224}, 2017{\natexlab{a}}.

\bibitem[Huang et~al.(2017{\natexlab{b}})Huang, Liu, Weinberger, and van~der
  Maaten]{huang2016densely}
Huang, G., Liu, Z., Weinberger, K.~Q., and van~der Maaten, L.
\newblock Densely connected convolutional networks.
\newblock \emph{CVPR}, 2017{\natexlab{b}}.

\bibitem[Ioffe \& Szegedy(2015)Ioffe and Szegedy]{ioffe2015batch}
Ioffe, S. and Szegedy, C.
\newblock Batch normalization: Accelerating deep network training by reducing
  internal covariate shift.
\newblock \emph{arXiv preprint arXiv:1502.03167}, 2015.

\bibitem[Kingma \& Ba(2014)Kingma and Ba]{kingma2014adam}
Kingma, D.~P. and Ba, J.
\newblock Adam: A method for stochastic optimization.
\newblock \emph{arXiv preprint arXiv:1412.6980}, 2014.

\bibitem[Krizhevsky \& Hinton(2009)Krizhevsky and
  Hinton]{krizhevsky2009learning}
Krizhevsky, A. and Hinton, G.
\newblock Learning multiple layers of features from tiny images.
\newblock 2009.

\bibitem[Liu et~al.(2017)Liu, Zoph, Shlens, Hua, Li, Fei-Fei, Yuille, Huang,
  and Murphy]{liu2017progressive}
Liu, C., Zoph, B., Shlens, J., Hua, W., Li, L.-J., Fei-Fei, L., Yuille, A.,
  Huang, J., and Murphy, K.
\newblock Progressive neural architecture search.
\newblock \emph{arXiv preprint arXiv:1712.00559}, 2017.

\bibitem[Liu et~al.(2018)Liu, Simonyan, Vinyals, Fernando, and
  Kavukcuoglu]{liu2017hierarchical}
Liu, H., Simonyan, K., Vinyals, O., Fernando, C., and Kavukcuoglu, K.
\newblock Hierarchical representations for efficient architecture search.
\newblock \emph{ICLR}, 2018.

\bibitem[Loshchilov \& Hutter(2016)Loshchilov and Hutter]{loshchilov2016sgdr}
Loshchilov, I. and Hutter, F.
\newblock Sgdr: stochastic gradient descent with restarts.
\newblock \emph{arXiv preprint arXiv:1608.03983}, 2016.

\bibitem[Mendoza et~al.(2016)Mendoza, Klein, Feurer, Springenberg, and
  Hutter]{mendoza2016towards}
Mendoza, H., Klein, A., Feurer, M., Springenberg, J.~T., and Hutter, F.
\newblock Towards automatically-tuned neural networks.
\newblock In \emph{Workshop on Automatic Machine Learning}, 2016.

\bibitem[Negrinho \& Gordon(2017)Negrinho and
  Gordon]{negrinho2017deeparchitect}
Negrinho, R. and Gordon, G.
\newblock Deeparchitect: Automatically designing and training deep
  architectures.
\newblock \emph{arXiv preprint arXiv:1704.08792}, 2017.

\bibitem[Pham et~al.(2018)Pham, Guan, Zoph, Le, and Dean]{pham2018efficient}
Pham, H., Guan, M.~Y., Zoph, B., Le, Q.~V., and Dean, J.
\newblock Efficient neural architecture search via parameter sharing.
\newblock \emph{arXiv preprint arXiv:1802.03268}, 2018.

\bibitem[Real et~al.(2017)Real, Moore, Selle, Saxena, Suematsu, Le, and
  Kurakin]{real2017large}
Real, E., Moore, S., Selle, A., Saxena, S., Suematsu, Y.~L., Le, Q., and
  Kurakin, A.
\newblock Large-scale evolution of image classifiers.
\newblock \emph{ICML}, 2017.

\bibitem[Real et~al.(2018)Real, Aggarwal, Huang, and Le]{real2018regularized}
Real, E., Aggarwal, A., Huang, Y., and Le, Q.~V.
\newblock Regularized evolution for image classifier architecture search.
\newblock \emph{arXiv preprint arXiv:1802.01548}, 2018.

\bibitem[Simonyan \& Zisserman(2014)Simonyan and Zisserman]{simonyan2014very}
Simonyan, K. and Zisserman, A.
\newblock Very deep convolutional networks for large-scale image recognition.
\newblock \emph{arXiv preprint arXiv:1409.1556}, 2014.

\bibitem[Szegedy et~al.(2015)Szegedy, Liu, Jia, Sermanet, Reed, Anguelov,
  Erhan, Vanhoucke, and Rabinovich]{szegedy2015going}
Szegedy, C., Liu, W., Jia, Y., Sermanet, P., Reed, S., Anguelov, D., Erhan, D.,
  Vanhoucke, V., and Rabinovich, A.
\newblock Going deeper with convolutions.
\newblock In \emph{CVPR}, 2015.

\bibitem[Szegedy et~al.(2016)Szegedy, Vanhoucke, Ioffe, Shlens, and
  Wojna]{szegedy2016rethinking}
Szegedy, C., Vanhoucke, V., Ioffe, S., Shlens, J., and Wojna, Z.
\newblock Rethinking the inception architecture for computer vision.
\newblock In \emph{CVPR}, 2016.

\bibitem[Szegedy et~al.(2017)Szegedy, Ioffe, Vanhoucke, and
  Alemi]{szegedy2017inception}
Szegedy, C., Ioffe, S., Vanhoucke, V., and Alemi, A.~A.
\newblock Inception-v4, inception-resnet and the impact of residual connections
  on learning.
\newblock In \emph{AAAI}, 2017.

\bibitem[Tai et~al.(2015)Tai, Socher, and Manning]{tai2015improved}
Tai, K.~S., Socher, R., and Manning, C.~D.
\newblock Improved semantic representations from tree-structured long
  short-term memory networks.
\newblock \emph{ACL}, 2015.

\bibitem[Veit et~al.(2016)Veit, Wilber, and Belongie]{veit2016residual}
Veit, A., Wilber, M.~J., and Belongie, S.
\newblock Residual networks behave like ensembles of relatively shallow
  networks.
\newblock In \emph{NIPS}, 2016.

\bibitem[Williams(1992)]{williams1992simple}
Williams, R.~J.
\newblock Simple statistical gradient-following algorithms for connectionist
  reinforcement learning.
\newblock In \emph{Reinforcement Learning}. 1992.

\bibitem[Xie et~al.(2017)Xie, Girshick, Doll{\'a}r, Tu, and
  He]{xie2017aggregated}
Xie, S., Girshick, R., Doll{\'a}r, P., Tu, Z., and He, K.
\newblock Aggregated residual transformations for deep neural networks.
\newblock In \emph{CVPR}, 2017.

\bibitem[Zhang et~al.(2017)Zhang, Zhou, Lin, and Sun]{zhang2017shufflenet}
Zhang, X., Zhou, X., Lin, M., and Sun, J.
\newblock Shufflenet: An extremely efficient convolutional neural network for
  mobile devices.
\newblock \emph{arXiv preprint arXiv:1707.01083}, 2017.

\bibitem[Zhong et~al.(2017)Zhong, Yan, and Liu]{zhong2017practical}
Zhong, Z., Yan, J., and Liu, C.-L.
\newblock Practical network blocks design with q-learning.
\newblock \emph{arXiv preprint arXiv:1708.05552}, 2017.

\bibitem[Zoph \& Le(2017)Zoph and Le]{zoph2016neural}
Zoph, B. and Le, Q.~V.
\newblock Neural architecture search with reinforcement learning.
\newblock \emph{ICLR}, 2017.

\bibitem[Zoph et~al.(2017)Zoph, Vasudevan, Shlens, and Le]{zoph2017learning}
Zoph, B., Vasudevan, V., Shlens, J., and Le, Q.~V.
\newblock Learning transferable architectures for scalable image recognition.
\newblock \emph{arXiv preprint arXiv:1707.07012}, 2017.

\end{thebibliography}
\bibliographystyle{icml2018}

\clearpage
\appendix
%\onecolumn

\section{Architecture Search Starting from Scratch}
Beside utilizing state-of-the-art human-designed architectures, we also perform architecture search starting from scratch (i.e. a chain of identity mappings) to learn how much we can benefit from reusing existing well-designed architectures. The structure of the start point is provided in Table~\ref{tab:identity_start_point}, where the identity mappings are later replaced by sampled cells to get new architectures and all other configurations keep the same as the ones used in Section~\ref{sec:exp}. 

\begin{figure}[h]
	\centering
	\includegraphics[width=\columnwidth]{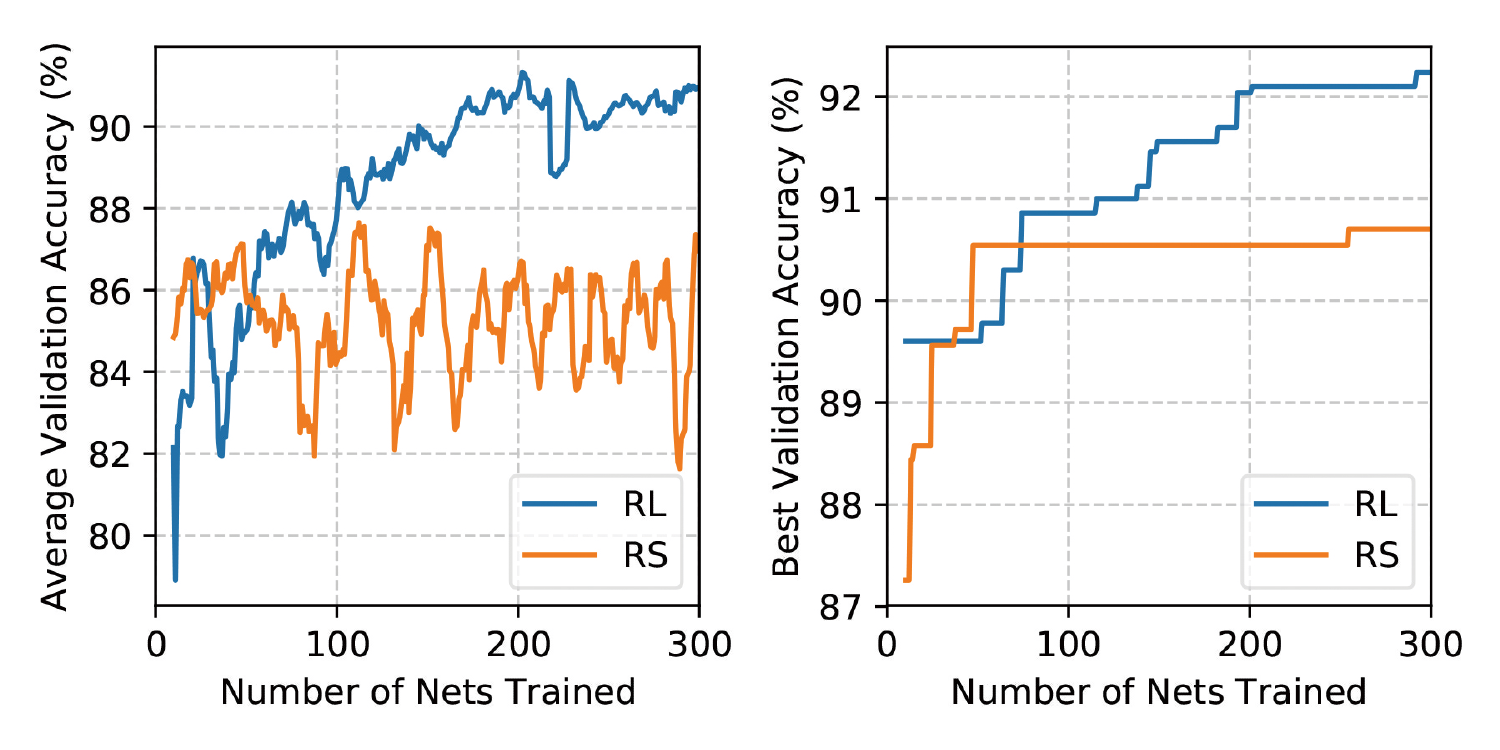}
	\caption{Progress of the architecture search process starting from scratch (a chain of identity maps) on CIFAR-10.}
	\label{fig:arch_search_progress_identity}
\end{figure}

The progress of the architecture search process is reported in Figure~\ref{fig:arch_search_progress_identity}, where we can observe similar trends as the ones in Figure~\ref{fig:arch_search_progress}. Moreover, we find that the advantage of RL over RS is larger in this case (RL achieves 1.54\% better validation accuracy than RS). After 300 epochs training on CIFAR-10, the best RL identified cell reaches 3.93\% test error with 11.5M parameters, which is better than 4.44\% given by the best random cell with 10.0M parameters, but is far worse than 3.14\% given by TreeCell-A with 5.7M parameters.  

\begin{table}[h]
	\centering
	\caption{Start point network with identity mappings on CIFAR-10.}\label{tab:identity_start_point}
	\vspace{5pt}
	\resizebox{\columnwidth}{!}{  
		\begin{tabular}{c | c | c }
		\hline
        Model architecture & Feature map size & Output channels\\
        \hline
        $3 \times 3$ Conv & $32 \times 32$ & 48 \\
        \hline
        [identity mapping] $\times 4$ & $32 \times 32$ & 48 \\
        \hline
        $1 \times 1$ Conv & $32 \times 32$ & 96 \\
        $3 \times 3$ average pooling, stride 2 & $16 \times 16$ & 96 \\
        \hline
        [identity mapping] $\times 4$ & $16 \times 16$ & 96 \\
        \hline
        $1 \times 1$ Conv & $16 \times 16$ & 192 \\
        $3 \times 3$ average pooling, stride 2 & $8 \times 8$ & 192 \\
        \hline
        [identity mapping] $\times 4$ & $8 \times 8$ & 192 \\
        \hline
        $8 \times 8$ global average pooling & $1 \times 1$ & 192 \\
        \hline
        10-dim fully-connected, softmax & & \\
		\end{tabular}
	}
\end{table}

\section{Details of Architecture Space}
We find the following 2 tricks effective for reaching good performances with the tree-structured architecture space in our experiments. 

\subsection{Group Convolution} 
The base networks (i.e. DenseNets and PyramidNets) in our experiments use standard $3 \times 3$ group convolution instead of normal $3 \times 3$ convolution and the number of groups $G$ is chosen from $\{1, 2, 4\}$ according to the sampled tree-structured cell. Specifically, if the merge scheme of the root node is $concatenation$, $G$ is set to be 1; if the merge scheme is $add$ and the number of branches is 2, $G$ is set to be 2; if the merge scheme is $add$ and the number of branches is 3, $G$ is set to be 4. As such, we can make different sampled cells have a similar number of parameters as the normal $3 \times 3$ convolution layer. 

\subsection{Skip Node Connection and BN layer}
Inspired by PyramidNets \cite{han2016deep} that add an additional batch normalization (BN) \cite{ioffe2015batch} layer at the end of each residual unit, which can enable the network to determine whether the corresponding residual unit is useful and has shown to improve the capacity of the network architecture. Analogously, in a tree-structured cell, we insert a skip connection for each child (denoted as $N^c_i (\cdot)$) of the root node, and merge the outputs of the child node and its corresponding skip connection via $add$. Additional, the output of the child node goes through a BN layer before it is merged. As such the output of the child node $N^c_i (\cdot)$ with input feature map $\bx$ is given as:
\begin{equation}
    O_i = add(\bx, BN(N^c_i(\bx))).
\end{equation}
In this way, intuitively, each unit with tree-structured cell can at least go back to the original unit if the cell is not helpful here. 

\section{Detailed Structure of TreeCell-B}

\begin{figure}[h]
	\centering
	\includegraphics[width=0.8\columnwidth]{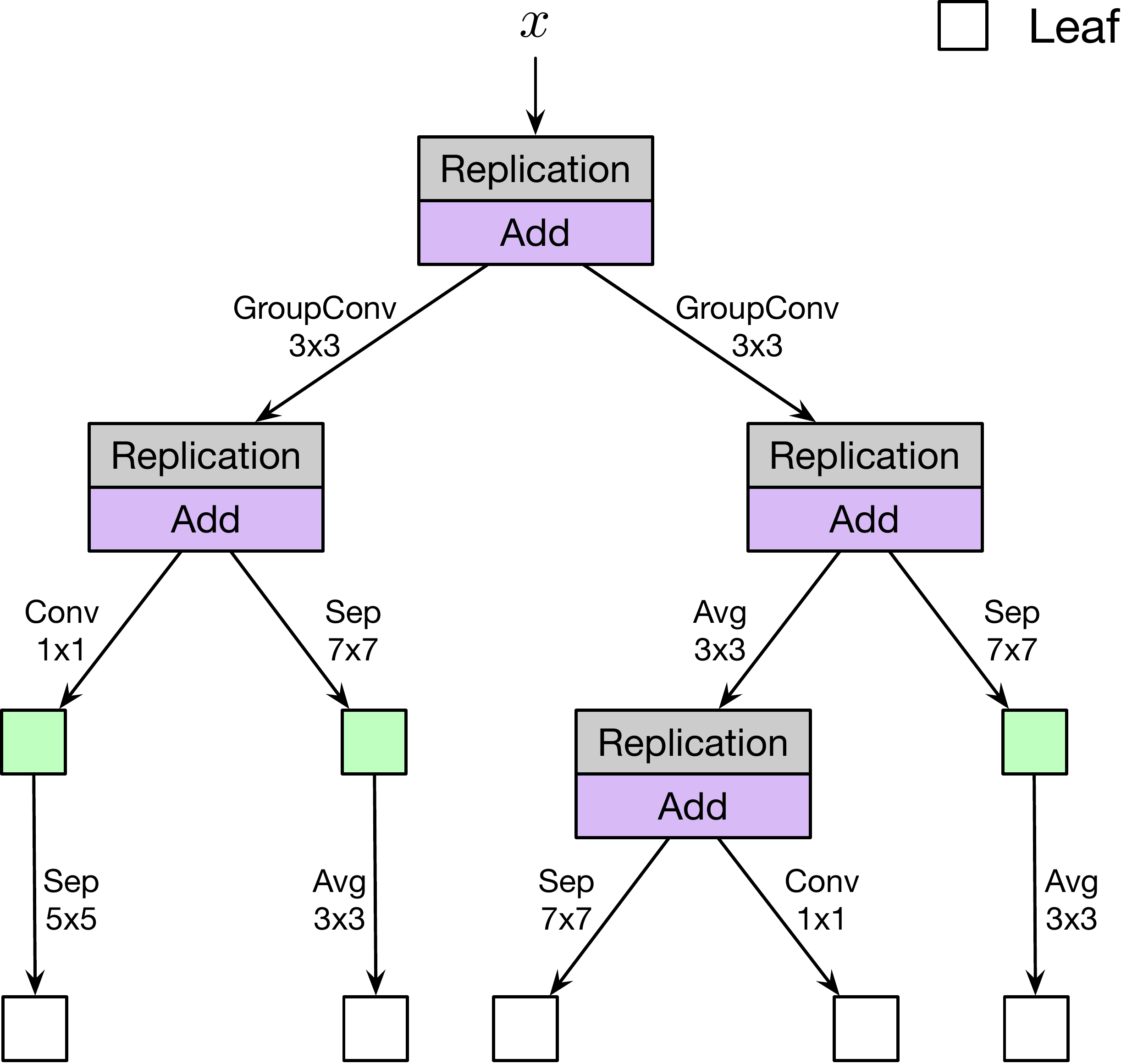}
	\caption{Detailed structure of TreeCell-B.}
	\label{fig:tree-cell-b}
\end{figure}

\section{Meta-Controller Training Procedure}
\begin{algorithm}[h]
	\caption{Path-Level Efficient Architecture Search}
	\label{alg:path-level-eas}
	{\small	
		\begin{algorithmic}[1]
			\REQUIRE
			base network $baseNet$, training set $trainSet$, validation set $valSet$, batch size $B$, maximum number of networks $M$ 
			\STATE $trained = 0$  // Number of trained networks
            \STATE $P_{nets} = []$ // Store results of trained networks
            \STATE randomly initialize the meta-controller $C$
            \STATE $G_c = []$ // Store gradients to be applied to $C$
            \WHILE{$trained < M$}
            \STATE meta-controller $C$ samples a tree-structured $cell$
            \IF{$cell$ in $P_{nets}$}
            \STATE get the validation accuracy $acc_v$ of $cell$ from $P_{nets}$
            \ELSE 
            \STATE model = train(trans($baseNet$, $cell$), $trainSet$)
            \STATE $acc_v$ = evel(model, $valSet$)
            \STATE add ($cell$, $acc_v$) to $P_{nets}$
            \STATE $trained = trained + 1$ 
            \ENDIF
            \STATE compute gradients according to ($cell$, $acc_v$) and add to $G_c$
            \IF{$len(G_c) == B$}
            \STATE update $C$ according to $G_c$
            \STATE $G_c = []$
            \ENDIF
            \ENDWHILE	
		\end{algorithmic}
	}
\end{algorithm}

\end{document}